%% file: template.tex
\documentclass{Interspeech}

 \interspeechcameraready

\title{Aligning ASR Evaluation with Human and LLM Judgments: Intelligibility Metrics Using Phonetic, Semantic, and NLI Approaches}

\author{Bornali}{Phukon}
\author{Xiuwen}{Zheng}
\author{Mark}{Hasegawa-Johnson}

\affiliation[nocounter]{Department of Electrical and Computer Engineering}{University of Illinois Urbana-Champaign}{USA}

\email{bornalip@illinois.edu, xiuwenz2@illinois.edu, jhasegaw@illinois.edu}


\usepackage{comment}

\begin{document}

\maketitle

\begin{abstract}

    Traditional ASR metrics like WER and CER fail to capture intelligibility, especially for dysarthric and dysphonic speech, where semantic alignment matters more than exact word matches. ASR systems struggle with these speech types, often producing errors like phoneme repetitions and imprecise consonants, yet the meaning remains clear to human listeners. We identify two key challenges: (1) Existing metrics do not adequately reflect intelligibility, and (2) while LLMs can refine ASR output, their effectiveness in correcting ASR transcripts of dysarthric speech remains underexplored. To address this, we propose a novel metric integrating Natural Language Inference (NLI) scores, semantic similarity, and phonetic similarity. Our ASR evaluation metric achieves a 0.890 correlation with human judgments on Speech Accessibility Project data, surpassing traditional methods and emphasizing the need to prioritize intelligibility over error-based measures.
\end{abstract}

\input{section1}

\input{section2}
\input{section3}

\input{section4}

\input{section5}

\bibliographystyle{IEEEtran}
\bibliography{mybib.bib}

\end{document}

%% file: section1.tex
\section{Introduction}
\label{sec:intro}

Traditional Automatic Speech Recognition (ASR) evaluation metrics such as Word Error Rate (WER) and Character Error Rate (CER) fail to capture intelligibility, particularly for dysarthric and dysphonic speech, where semantic alignment matters more than exact word or character matches. ASR systems struggle with these speech types due to deviations from typical speech patterns and limited relevant training data [1]. As a result, transcripts often contain errors such as repeated phonemes, breathy interjections, and imprecise consonants, common characteristics of dysarthric and dysphonic speech. Despite these issues, the intended meaning often remains clear to human listeners. 

For example, as illustrated in Figure \ref{fig:example_fig}, Example 1 shows an ASR hypothesis with phoneme repetitions, resulting in an extremely high WER of 255.56, yet it remains intelligible and conveys the correct meaning. In Example 2, a hypothesis with imprecise consonants had a WER of 30.77\%, but after processing with GPT-4, the output perfectly aligned with the reference. These cases highlight how traditional metrics can severely penalize transcripts that remain intelligible, emphasizing the need for a more nuanced evaluation. The ability of LLMs like GPT4 to refine ASR output further underscores the limitations of WER in assessing intelligibility.
\input{Image/example}

Two key challenges emerge from these observations. First, existing ASR metrics do not sufficiently capture intelligibility. Alternative metrics such as \textit{BERTScore} \cite{zhang2020bert} generally offer semantic similarity but may fail to detect logical contradictions in certain cases. For example, it assigns a high similarity score (0.8741) between the following contradictory sentences:
\begin{quote} \textit{Hypothesis: "The project deadline is flexible"} vs. \textit{Reference: "The project deadline is not flexible"}. \end{quote}
This limitation suggests that semantic similarity alone is insufficient for ASR evaluation. 

Second, while LLMs can improve ASR outputs, their correction capability remains underexplored. Recent advancements suggest that LLMs can refine ASR-generated transcripts\cite{guo2019spelling}, \cite{shivakumar2019learning}, \cite{tanaka2018neural}, \cite{gu2024denoising}, correcting phonetic errors and improving intelligibility. However, it is unclear which types of errors LLMs can correct most effectively and how this aligns with human perception. Understanding the correctability of ASR transcripts can inform better ASR evaluation methods.

While WER, phonetic similarity, and semantic similarity capture different aspects of ASR quality, they do not directly indicate how intelligible an ASR output is or how easily it can be corrected. Our primary objective is to develop a metric for intelligibility, but given that intelligibility and correctability are closely related, we investigate whether factors influencing correctability can be integrated into our metric. 

To better assess ASR intelligibility, we introduce a new metric that combines Natural Language Inference (NLI) score, semantic similarity, and phonetic similarity with optimized weights.
While phonetic similarity, semantic similarity, and logical entailment (NLI) have been studied individually, our novelty lies in combining these three elements into a unified metric specifically designed for evaluating ASR intelligibility. This integrated approach better reflects human perception and ASR correctability.  

Using data from the Speech Accessibility Project (SAP), we analyze multiple metrics while assessing several ASR systems across varying severity levels of dysarthric speech. Leveraging insights from this analysis, our optimized metric achieves a high correlation (0.890) with human judgments, notably outperforming traditional metrics. The evaluation scripts are available at the GitHub repository\footnote{\url{https://github.com/BornaliP/SemScore}}.

%% file: Image/example.tex
\begin{figure}[t] %
    \centering
    \includegraphics[width=\columnwidth]{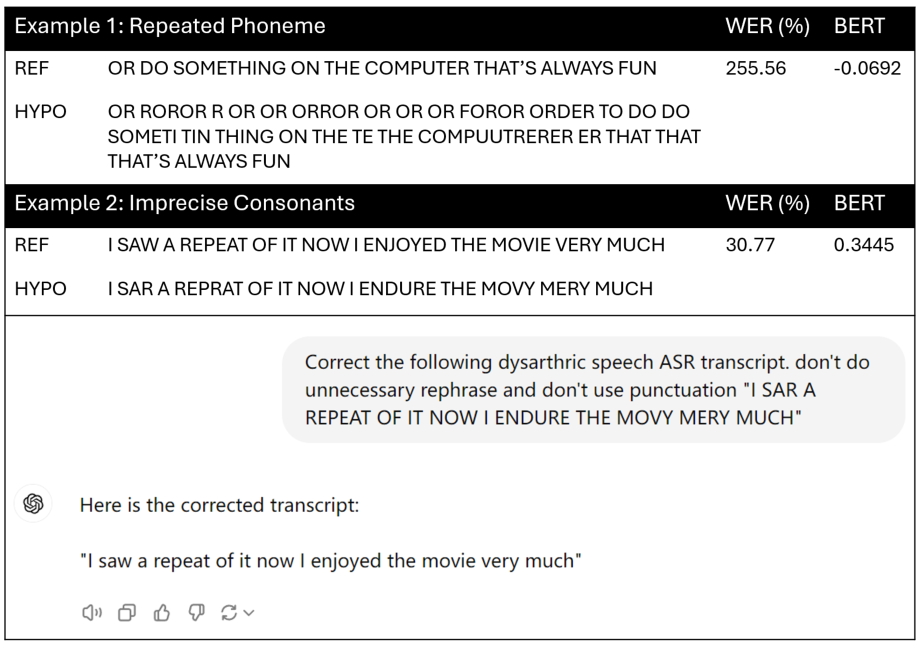} 
    \vspace{-10pt}
    \caption{Examples of dysarthric speech ASR Errors and GPT Corrections}
    
   \label{fig:example_fig} 
   \vspace{-2.0em}
\end{figure}

%% file: section2.tex
\section{Challenges in ASR evaluation for dysarthric speech }
\label{sec:dysartric_challanges}
Section \ref{sec:intro} highlighted the limitations of traditional ASR metrics for dysarthric speech evaluation. Here, we assess various ASR systems on dysarthric speech using these metrics, investigating whether a single metric can offer a comprehensive evaluation and identifying key components for effective assessment.

We evaluate multiple ASR systems, including \textit{Wav2vec 2.0}~\cite{baevski2020wav2vec} and \textit{Whisper}~\cite{radford2022robust}, with dysarthric speech from the validation split (23.32h) of SAP~\cite{omalley2023uiuc} release 2023-10-05 for this investigation. The experiments are conducted solely on SAP, as it is currently the largest and most diverse publicly available corpus for dysarthric speech, approximately ten times larger than any other comparable dataset, making it the most reliable benchmark for evaluating ASR performance on impaired speech. \textit{Wav2vec 2.0 base} is a transformer-based speech encoder pre-trained with Librispeech, which performs well in ASR by fine-tuning on a specific corpus to optimize the CTC~\cite{graves2006connectionist} loss. This work utilized two official wav2vec base models, fine-tuned on typical speech from $100$ hours and $960$ hours of Librispeech, respectively, as well as the best checkpoint trained by~\cite{Zheng2024}, using $151.47$ hours of dysarthric speech in the SAP-1005 training split. We refer to these models as wav2vec-(libri100, libri960 and sap1005) throughout the paper. \textit{Whisper base English} model utilized in this work is a multi-task multi-lingual transformer-based encoder-decoder model trained with 680,000 hours of audio using weak supervision, which significantly enhances its accuracy and robustness in ASR across different scenarios.

We evaluate the ASR systems using three metric categories: word-level, phonetic, and semantic similarity. For word-level similarity, we use WER, calculating it for each utterance and averaging the results to maintain consistency. Phonetic similarity is assessed through  \href{https://github.com/Kyubyong/g2p}{g2p}\footnote{\url{https://github.com/Kyubyong/g2p}} phoneme-level similarity and {Soundex code-based similarity}\footnote{\url{https://ics.uci.edu/~dan/genealogy/Miller/javascrp/soundex.htm}}. For semantic similarity, we apply BERTScore F1~\cite{zhang2020bert} and BLEURT~\cite{sellam2020bleurt}. Additionally, we use Heval~\cite{sasindran2023h}, a hybrid metric developed specifically for ASR systems, to examine how these metrics perform with dysarthric speech.

Table \ref{tab:eval_trad} presents the performance of the ASR systems across various metrics. The \textit{wav2vec-sap1005} model leads in WER (23.26\%) and phonetic similarity, achieving scores of 0.9294 (g2p) and 0.9748 (Soundex). This is due to its fine-tuning on dysarthric speech, enabling better handling of dysarthric speech. However, \textit{Whisper} excels in semantic similarity and the hybrid metric, likely benefiting from its training on a larger, more diverse dataset that includes semantic tasks like translation and multi-task learning. The results highlight that low WER does not always correspond to high semantic similarity, nor does high semantic similarity imply low WER, indicating that a single metric is insufficient for assessing both word-level accuracy and semantic correctness. We further examine how this insight holds across different severity levels.
\input{tab1}

\vspace{-2pt}

Table~\ref{tab:asr_results} shows the ASR performance across four severity levels of dysarthric speech: high (2h), medium (4h), low (7h), and very low (8h), based on \cite{Zheng2024}. In this table, we consider only Soundex (Psim II) for phonetic similarity, as both g2p and Soundex provide nearly the same interpretation, with Soundex offering better insight into phonetic variations.
\textit{Wav2vec-sap1005} outperforms \textit{Whisper} at high and medium severity levels, with WER differences of 16.89\% (51.50\% vs. 68.39\%) and 9.45\% (29.84\% vs. 39.29\%). However, \textit{Whisper} performs better at low and very low severity levels, with advantages of 1.61\% (19.64\% vs. 21.25\%) and 4.02\% (10.72\% vs. 14.74\%). This suggests \textit{Whisper} generalizes better to low-severity speech, while \textit{wav2vec-sap1005} excels in severe cases. Consistent with Table \ref{tab:eval_trad}, \textit{Whisper} achieves the highest semantic similarity across most severity levels, except for high severity.

\input{tab2}
\input{tab3}

Across all severity levels, \textit{wav2vec-sap1005} generates hypotheses with high phonetic similarity to the reference transcript. This is due to its exposure to the phonetic characteristics of dysarthric speech, allowing it to better capture the speech sounds of dysarthria: even when it does not have the lowest WER, its word errors tend to replace reference words with similar-sounding hypotheses. In contrast, \textit{Whisper} generates hypotheses with higher lexical semantic similarity (BERT and BLEURT) to the reference transcript but not necessarily with high phonetic similarity. The phonetic score differences between \textit{wav2vec-sap1005} and \textit{Whisper} grow as the severity increases. At high severity, the difference is 0.0609 (0.9135 for \textit{wav2vec-sap1005} vs. 0.8526 for \textit{whisper}). This gap narrows to 0.0356 at mid-severity (0.9664 vs. 0.9308). In low severity, the difference is further reduced to 0.0207 (0.9773 vs. 0.9566), and in very low severity, the difference becomes almost negligible at 0.0088 (0.9919 vs. 0.9831). 

Table \ref{tab:asr_results}  suggests that 
WER, lexical semantic similarity, and phonetic similarity measure slightly different aspects of the relationship between the reference and hypothesis transcripts but give little indication of the relative utility of each of these measures in judging the quality of an ASR hypothesis.  In order to determine the optimal combination of these disparate measures, it is necessary to consider the way in which an ASR hypothesis will be interpreted, either directly by a human reader or by a downstream natural language processing system.  The next two sections consider, first, a downstream LLM, and second, interpretation by a human reader.

%% file: tab1.tex
    \begin{table}[t!]
    \caption{\textbf{Evaluation of ASR Systems on Dysarthric Speech using WER, Phonetic, and Semantic Metrics.} ``Psim'' refers to ``phonetic similarity'', calculated by Levenshtein Distance (Psim I) and Soundex Algorithm (Psim II). The best entry in each column is in boldface.}
    \label{tab:asr_results}
    \renewcommand{\arraystretch}{1.3}
    \scalebox{0.82}{
    \begin{tabular}{|l|c|cc|ccc|}
    \hline
    \textbf{Model}  &\textbf{WER\%}   & \textbf{Psim I} & \textbf{Psim II} & \textbf{BERT} & \textbf{Bleurt}  & \textbf{Heval}\\
    \hline
    Wav2vec &   & & &   &  & \\
    
    \hspace{0.2cm}-libri100  & 40.27   & 0.8728 & 0.9247 & 0.4903  & -0.6121 & 0.2484\\
    \hspace{0.2cm}-libri960 & 33.15  & 0.8931 & 0.9343 & 0.5781  & -0.3756 & 0.1893\\
    \hspace{0.2cm}-sap1005 & \textbf{23.26}  & \textbf{0.9294} & \textbf{0.9748} & 0.6896  & -0.1262 & 0.1031\\
    \hline
    Whisper   & 24.59  & 0.9269 & 0.9519 & \textbf{0.7387}  & \textbf{0.2098} & \textbf{0.1005}\\
    \hline
    \end{tabular}
    }
    \label{tab:eval_trad}
    \vspace{-1.5em}
    \end{table}

%% file: tab2.tex
\begin{table}[t!]
\caption{\textbf{Evaluation of ASR Systems on Dysarthric Speech, classified by speech impairment severity levels.}  The best entry in each block is in boldface.}
\renewcommand{\arraystretch}{1.3}
\scalebox{0.84}{
\begin{tabular}{|c|l|c|c|ccc|}
\hline
& \textbf{Model} & \textbf{WER\%}  & \textbf{Psim II} & \textbf{BERT}  & \textbf{Bleurt} & \textbf{Heval}\\
\hline
H & Wav2vec &&&&& \\
 & \hspace{0.2cm}-libri100 & 91.50  & 0.7618 & -0.1489 & -1.3061 & 0.9205  \\
 & \hspace{0.2cm}-libri960 & 82.49  & 0.7975 & -0.0577 & -1.3483 & 0.7597  \\
& \hspace{0.2cm}-sap1005  & \textbf{51.50} &  \textbf{0.9135} & \textbf{0.2969} & -0.9798 & \textbf{0.3661} \\ \cline{2-7}
& Whisper  & 68.39 & 0.8526 & 0.2370 & \textbf{-0.9212} & 0.4553  \\
\hline

M & Wav2vec &&&&& \\
 & \hspace{0.2cm}-libri100 & 56.39  & 0.9034 & 0.3128 & -1.09 & 0.3242  \\
 & \hspace{0.2cm}-libri960 & 46.38  & 0.9195 & 0.4306 & -0.8743 & 0.2420  \\
& \hspace{0.2cm}-sap1005  & \textbf{29.84}  &\textbf{0.9664} & 0.6094 & -0.4109 & \textbf{0.1269 } \\\cline{2-7}
& Whisper  & 39.29 &  0.9308 & \textbf{0.6347} & \textbf{-0.1155} & 0.1307 \\
\hline

L & Wav2vec &&&&& \\
 & \hspace{0.2cm}-libri100 & 36.98 &  0.9280 & 0.5190 & -0.6609 & 0.1903  \\
 & \hspace{0.2cm}-libri960 & 31.39  & 0.9314 & 0.5868 & -0.4622 & 0.1475  \\
& \hspace{0.2cm}-sap1005  & 21.25  & \textbf{0.9773} & 0.7117 & -0.0981 & 0.0761  \\\cline{2-7}
& Whisper  & \textbf{19.64 } & 0.9566 & \textbf{0.7744 } &\textbf{ 0.2687} & \textbf{0.0646}  \\
\hline
VL & Wav2vec &&&&& \\
 & \hspace{0.2cm}-libri100 & 22.39  & 0.9728 & 0.7124 & -0.1571 & 0.0952  \\
 & \hspace{0.2cm}-libri960 & 15.83  & 0.9781 & 0.8020 & 0.1920 & 0.0585  \\
& \hspace{0.2cm}-sap1005  & 14.74  & \textbf{0.9919} & 0.8077 & 0.2034 & 0.0498  \\\cline{2-7}
& Whisper  & \textbf{10.72} &0.9831 & \textbf{0.8836} & \textbf{0.6015} & \textbf{0.0291}  \\
\hline
\end{tabular}}
\label{tab:asr_results}
\end{table}

%% file: tab3.tex
\begin{table}[tbp]
\centering
\caption{\textbf{Evaluation of ASR Systems on Dysarthric Speech, w/ and w/o GPT corrections.} ``Wav2vec'' in this table refers to wav2vec-sap1005. The best entry in each block is in boldface.}
\renewcommand{\arraystretch}{1.3}
\scalebox{0.80}{
\begin{tabular}{|l|l|c|c|ccc|}
\hline
& \textbf{Model} & \textbf{WER\%} & \textbf{Psim II} & \textbf{BERT}  & \textbf{Bleurt} & \textbf{Heval} \\ 
\hline
w/o GPT & Wav2vec & \textbf{31.11} & \textbf{0.9752} & 0.5954 & -0.3284 & \textbf{0.1699}  \\
&Wavllm & 35.85 & 0.9718 & 0.5334 & -0.5769 & 0.1976  \\ 
&Whisper & 34.72 & 0.9387 & \textbf{0.6516} & \textbf{0.0183} & 0.1700  \\ 
\hline
w/ GPT & Wav2vec & \textbf{24.26} & \textbf{0.9673} & \textbf{0.7023} & \textbf{0.1893} & \textbf{0.1197}  \\ 
& Wavllm & 27.3 & 0.9655 & 0.6714 & 0.0839 & 0.1349  \\
& Whisper & 40.99 & 0.9198 & 0.6254 & 0.0265 & 0.1863  \\ 
\hline
w/ GPT, & Wav2vec& \textbf{21.91} & \textbf{0.9813} & \textbf{0.7234} & \textbf{0.2362} & \textbf{0.1084} \\ 
improved& Wavllm & 24.74 & 0.9782 & 0.6903 & 0.1216 & 0.1220  \\ 
only& Whisper & 29.31 & 0.9428 & 0.6909 & 0.1591 & 0.1537  \\ 
\hline
\end{tabular}
}
\label{tab:gpt_results}
\vspace{-1.5em}
\end{table}

%% file: section3.tex
\section{Evaluating ASR Correctability Using Large Language Models}
\label{sec:gpt_correct} 

While WER, phonetic similarity, and semantic similarity capture different aspects of ASR quality, they do not directly indicate how intelligible an ASR output is or how easily it can be corrected. 
Much recent research has demonstrated that LLM correction can reduce the WER of an ASR transcript~\cite{guo2019spelling,shivakumar2019learning,tanaka2018neural,gu2024denoising,tomanek2024large}, 
but the aspects of a transcript that make it correctable or uncorrectable have not been fully explored.
In this section, we evaluate whether phonetic similarity, semantic similarity, and WER contribute to predicting ASR correctability. To investigate this, we tested two LLM-based correction settings: (1) the WavLLM model \cite{hu2024wavllm}, which integrates LLMs with Whisper~\cite{radford2022robust} and WavLM~\cite{chen2022wavlm} for ASR tasks, and (2) the GPT-4 API \cite{achiam2023gpt}, using a prompt to generate corrected transcripts. GPT-4 is chosen for its reliability over LLaMA3 and GPT-3, and we limited the analysis to 1,441 samples from the SAP1005 dataset (1,321 short, 50 moderate, and 50 long samples), using a model temperature of  0.1. The transcript lengths are categorized by word-level tokens: short (fewer than 10), moderate (10 to 25), and long (more than 30).

{\par \textit{Prompt: "Correct this ASR transcript for readability, clarity, and spelling while preserving the original meaning. Make minimal changes, fixing errors or incorrect terms and names without unnecessary rephrasing."}

Table \ref{tab:gpt_results} presents the performance of the top two models from our previous experiment, \textit{wav2vec-sap1005} and \textit{Whisper}, along with \textit{WavLLM}, both before and after applying GPT-4 correction. 
Of these three, only \textit{wav2vec-sap1005} has been fine-tuned on dysarthric speech.
Results are presented in three categories: (1) without GPT correction, (2) with GPT correction averaged across all samples, and (3) with GPT correction only for transcripts where oracle analysis indicated that WER was improved by the correction, and without GPT correction if
the correction degraded WER. The third category is a type of oracle experiment, of interest because GPT improvements and degradations have very different character: when GPT improves a transcript, it makes small changes, but when it degrades a transcript, it often does so by generating a long hallucinated transcript with relatively little relationship to its input.
For instance, an example of GPT-4 correction successfully improving intelligibility can be seen in the case of the transcript:

\begin{itemize}
    \item \textbf{Hypothesis:} ``SET THE AIR CONDITIONING CONDITI CONDITIONING TO SEESEVENTY SE E EE SEENT EIGEIGHT''
    \item \textbf{Reference:} ``SET THE AIR CONDITION DITIONING TO SEV SEVENTY EIGHT''
    \item \textbf{GPT-Corrected:} ``SET THE AIR CONDITIONING CONDITIONING TO SEVENTY EIGHT''
\end{itemize}

Here, the GPT correction successfully reduces redundancy and aligns the transcript more closely with the ground truth, reducing WER from 1.0 to 0.333.

On the other hand, GPT-4 correction sometimes introduces significant hallucinations, leading to transcript degradation. An example of this phenomenon is:

\begin{itemize}
    \item \textbf{Hypothesis:} ``OPEN GULAMNBA''
    \item \textbf{Reference:} ``OPEN DUOLINGO''
    \item \textbf{GPT-Corrected:} ``OPEN GULAMNBA CORRECTED TEXT OPEN GYM NBA''
\end{itemize}

In this case, GPT-4 completely misinterprets the intended phrase "DUOLINGO" and introduces an unrelated correction, resulting in an even worse transcription. The WER increases significantly from 0.5 to 3.0, demonstrating a severe case of hallucination.

Table \ref{tab:gpt_results}} shows that \textit{WavLLM} underperforms in WER compared to \textit{Whisper} and \textit{wav2vec-sap1005}, but surpasses \textit{Whisper} in phonetic similarity. Among the models, \textit{wav2vec-sap1005} showed the most improvement after GPT correction, with WER dropping from 31.11 to 24.26 (6.85\%) and to 21.91 (9.2\%) for improved cases. BERTScore improved by 0.1069, BLEURT by 0.5177, and HEval by 0.0502, though phonetic similarity remained unchanged, 
suggesting that GPT-4 improved semantic scores by finding a phonetically similar word sequence with a higher LLM 
probability. This highlights \textit{wav2vec-sap1005} as highly compatible with GPT-based corrections across multiple metrics.  Similarly, \textit{WavLLM} showed improvements with WER dropping from 35.85 to 27.23 (8.62\%), BERTScore improving by 0.138, BLEURT by 0.6608, and HEval by 0.0627. 
\textit{WavLLM} performed well in phonetic similarity but underperformed in all other metrics, yet \textit{WavLLM} has higher correctability than other models, suggesting the possibility that phonetic similarity has some influence on correctability. Across all waveform files produced by all three systems, the correctability of an individual transcript (the difference between WER with versus without correction) is significantly correlated with the phonetic similarity score of the uncorrected transcript, with a significance level of $p = 0.0038$, but with a small effect size (correlation coeficient = 0.0795), suggesting that other factors also play a role. 

\textit{Whisper} behaved differently than either \textit{wav2vec-sap1005} or \textit{WavLLM:} GPT-4 corrections did not improve \textit{Whisper} transcripts on average (except for a small BLEURT gain), though the oracle setting showed improvement (when failed corrections were discarded). This suggests that \textit{Whisper} outputs have fewer correctable errors. However, its highest BERTScore indicates stronger semantic correctness, supporting the idea that 
improved semantic accuracy of a baseline ASR transcript does not contribute to its LLM correctability.

%% file: section4.tex
\section{A new integrated metric for dysartic speech ASR evaluation }
\label{sec:new_metric}
\input{Image/metric_corr}

Sections \ref{sec:intro}, \ref{sec:dysartric_challanges}, and \ref{sec:gpt_correct} highlight that: 1) A single metric is insufficient for evaluating dysarthric speech ASR systems; WER captures word accuracy but misses semantic aspects. 2) Metrics like BERTScore measure semantic similarity but do not 
predict LLM correctability.
3) Phonetic similarity 
is significantly correlated with LLM correctability, but the correlation coefficient is not high enough 
to support the claim that phonetic similarity is the only predictor of correctability.

Motivated by these insights, we introduce a new metric that integrates semantic similarity, 
phonetic similarity,
and a new third component: natural language inference (NLI)
for a more comprehensive ASR evaluation.  To assess its effectiveness, six human annotators rated 100 ASR hypothesis-reference pairs on a scale of 1 to 5, evaluating intelligibility and how well the intended meaning is preserved. Each annotator rated the same 100 pairs.
 The correlation between annotators ranged from 0.78 to 0.93, indicating strong agreement among their ratings. The standard deviation of the individual rating scores was 1.14, reflecting reliable yet slightly varied judgments.


\par \textbf{Calculating  {NLI\_Score}:} 
An ASR hypothesis might be considered to correctly represent user intent if it can be proven to logically entails the reference transcript, even if the semantic and phonetic similarities between reference and hypothesis are quite low.
Entailment probabilities computed by a trained NLI model have been used to estimate the quality of machine translation and document summarization~\cite{chen2023menli} by computing a weighted linear average of the probability that the hypothesis entails the reference (in the notation of \cite{chen2023menli}: $P_e(\text{ref}\leftarrow\text{hypo})$), the probability that the hypothesis entails the reference ($P_e(\text{hypo}\leftarrow\text{ref})$),
the probability of mutual entailment ($P_e(\text{hypo}\leftrightarrow\text{ref})$),
the probabilities of all three types of contradiction ($P_c$), and the probability that there is neither entailment
nor contradiction ($P_n$).
Of these, the scalar probability $P_e(\text{hypo}\leftrightarrow\text{ref})$ was demonstrated to be the most effective metric of machine translation and document summarization quality~\cite{chen2023menli}, and we therefore adopt the same metric as our NLI score ($S_{NLI}$) of ASR hypothesis quality.
$S_{NLI}$ is set equal to the probability $P_e(\text{hypo}\leftrightarrow\text{ref})$ computed by a RoBERTa-large model \cite{liu2019roberta} that has beeen fine-tuned on the SNLI \cite{bowman-etal-2015-large}, MNLI, FEVER \cite{nie2019combining}, and ANLI \cite{nie2019adversarial} datasets.
To calculate the semantic score $S_{S}$, we used BERTScore, and for phonetic similarity $S_P$, we applied Soundex with Jaro-Winkler similarity, as discussed in Section \ref{sec:dysartric_challanges}. The integrated metric is defined as:
\[
\text{Integrated\_Score} = \alpha S_{NLI} + \beta S_S + \gamma S_P
\]
where \( \alpha \), \( \beta \), and \( \gamma \) represent the weights. Optimal weights were determined via linear regression using human ratings as the target.
We used 5-fold cross-validation to avoid overfitting, training the model on 80\% of the data and testing it on the remaining 20\%. 
Figure \ref{fig:example} shows that the integrated metric, using optimal weights, achieved the highest correlation (0.890) with human ratings, outperforming individual metrics. This demonstrates that the integrated approach better aligns with human judgment and is effective in evaluating ASR outputs with a focus on correctability.

After completion of cross-validation, weights were recomputed using the entire linear regression dataset (100 transcript pairs, each labeled by the average of intelligibility scores assigned by six human annotators).
The final normalized weights were: 
\[
\alpha = 0.40,  \beta =0.28,   \gamma =0.32
\]
The p-values for each coefficient were all found to be statistically significant (NLI Score: p = 1.47e-07, BERT Score: p = 0.0040, Phonetic Score: p = 0.0002). Mean Squared Error is 0.237, suggesting a reasonable fit of the model to the data. We also conducted a Shapiro-Wilk test for the normality of residuals, which returned a p-value of 0.9155, indicating that the residuals are normally distributed and satisfying an important assumption for linear regression. The normalized weights show that phonetic similarity (\(\gamma = 0.32\)) plays a larger role than semantic similarity (\(\beta = 0.28\)) in determining correctability, particularly for dysarthric speech. The NLI score has the highest weight (\(\alpha = 0.40\)), underscoring its 
ability to measure
logical consistency and coherence in ASR outputs.

%% file: Image/metric_corr.tex
\begin{figure}[ht!] %
    \centering
    \includegraphics[width=\columnwidth]{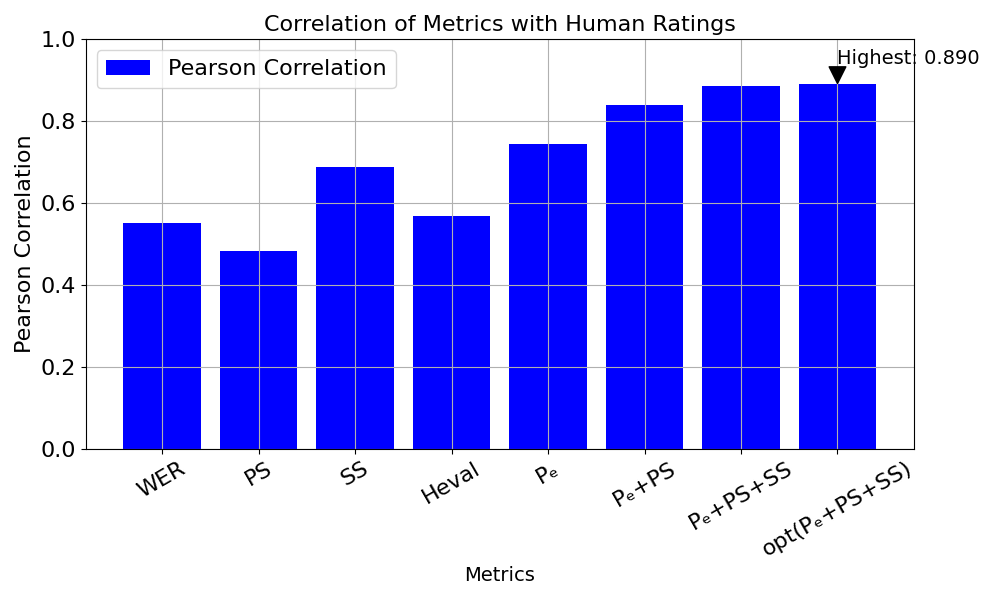} 
    \caption{Correlation of WER, Phonetic, Semantic, NLI, and Proposed Metrics with Human Ratings}
   \label{fig:example} 
\end{figure}

%% file: section5.tex
\section{Conclusion}

Traditional ASR evaluation metrics like WER and CER fail to capture intelligibility, particularly for dysarthric and dysphonic speech, where semantic alignment is more important than exact word matches. Our analysis highlights the limitations of existing metrics and demonstrates the potential of LLMs in refining ASR outputs. To address these gaps, we introduce a novel metric that integrates Natural Language Inference (NLI), semantic similarity, and phonetic similarity, offering a more comprehensive measure of ASR intelligibility.

By evaluating ASR systems on Speech Accessibility Project (SAP) data, our metric achieves a (0.890)  correlation with human judgments, significantly outperforming traditional approaches. These findings emphasize the need for ASR evaluation to prioritize intelligibility and correctability over rigid error-based measures, helping to develop more effective and user-centered ASR assessment methods.

\section{Acknowledgment}
This research utilized the Delta advanced computing and data resource, which is supported by the National Science Foundation (award OAC 2005572) and the State of Illinois. Delta is a joint effort of the University of Illinois Urbana-Champaign and its National Center for Supercomputing Applications.\footnote{\textit{Delta: An AI-Centric System for Scientific Discovery}. University of Illinois at Urbana-Champaign. \url{https://hdl.handle.net/2142/117179}}. The data used for this research was made possible by a grant to the University of Illinois from the AI Accessibility Coalition, whose members include Amazon, Apple, Google, Meta,
and Microsoft